\documentclass[letterpaper,10pt,conference]{ieeeconf}
\IEEEoverridecommandlockouts
\title{Stabilizing Fast Dynamical Systems Using Time-of-Flight Cameras}

\author{Anthony Czubarow, Antonio Terpin, and Raffaello D'Andrea%
\thanks{All authors are with the Institute for Dynamic Systems and Control,
ETH Z\"urich, Z\"urich, Switzerland.}%
}

\usepackage{cite}
\usepackage{paper-support}

\crefname{figure}{Fig.}{Figs.}
\Crefname{figure}{Fig.}{Figs.}

\begin{document}

\maketitle
\thispagestyle{empty}
\pagestyle{empty}

\begin{abstract}
Time-of-flight cameras are popular in robotics because they provide direct depth information and are compact, inexpensive, and less dependent on ambient visible light.
However, their low spatial resolution and depth noise raise questions about their suitability for stabilizing fast, unstable dynamics.
In this paper, we show that an inexpensive, low-resolution time-of-flight camera can provide angle feedback for reliable and precise balancing of an inverted pendulum on a cart, a canonical control benchmark.
These results open the door to using compact, inexpensive, low-resolution time-of-flight cameras for precise feedback control of fast, unstable systems.
\end{abstract}

\glsresetall
\section{Introduction}
\label{sec:introduction}
\Gls*{tof} cameras have become a practical sensing choice across robotics \cite{grzegorzek2013tof,hansard2013tof, may2009robust} because they deliver direct depth at video rates from a compact, low-power, and inexpensive package. Unlike passive RGB cameras, \gls*{tof} actively illuminates the scene, making it robust to low-texture, low-light, or repetitive-pattern environments \cite{alenya2014tof}.
\Gls*{tof} cameras have been used to acquire scene geometry for grasp and manipulation planning \cite{xue2012autonomous}.
They have also supplied visual feedback for human-guided robot motion \cite{marshall2012uncalibrated}, target tracking \cite{reiser2007using}, and robot positioning \cite{pomares2010visual}.

Our focus is \emph{precise stabilization of fast, unstable dynamics}.
Stabilizing an unstable equilibrium requires accurate camera feedback delivered quickly enough for the controller to counteract growing deviations.
Sensing limitations that may affect stabilization include systematic errors (e.g., depth distortion, integration-time and amplitude-related effects) and non-systematic effects (e.g., multipath/scattering, motion blur, and limited signal-to-noise ratio) \cite{foix2011lock,gudmundsson2007environmental}.

\begin{center}
    \textit{Can an inexpensive, low-resolution \gls*{tof} camera provide sufficiently accurate and timely angle feedback to stabilize an open-loop unstable equilibrium of a system with fast dynamics?}
\end{center}
In this work, we address this question by providing a \emph{proof of existence}: we stabilize an inverted pendulum on a cart using angle measurements from an inexpensive, low-resolution \gls*{tof} camera placed below the pendulum.
The motor encoder supplies cart-position measurements.
\reviewchange{We use \gls*{lqg} control, combining a \gls*{lqr} with a \gls*{kf}.}
This problem has long served as a canonical benchmark in control theory\cite{astrom2000swing,franklin2010feedback,tedrake2023underactuated,wei1995nonlinear,Zielinska2021,chung1995nonlinearpendulum}.
\begin{figure}
    \centering
    \includegraphics[width=.95\linewidth]{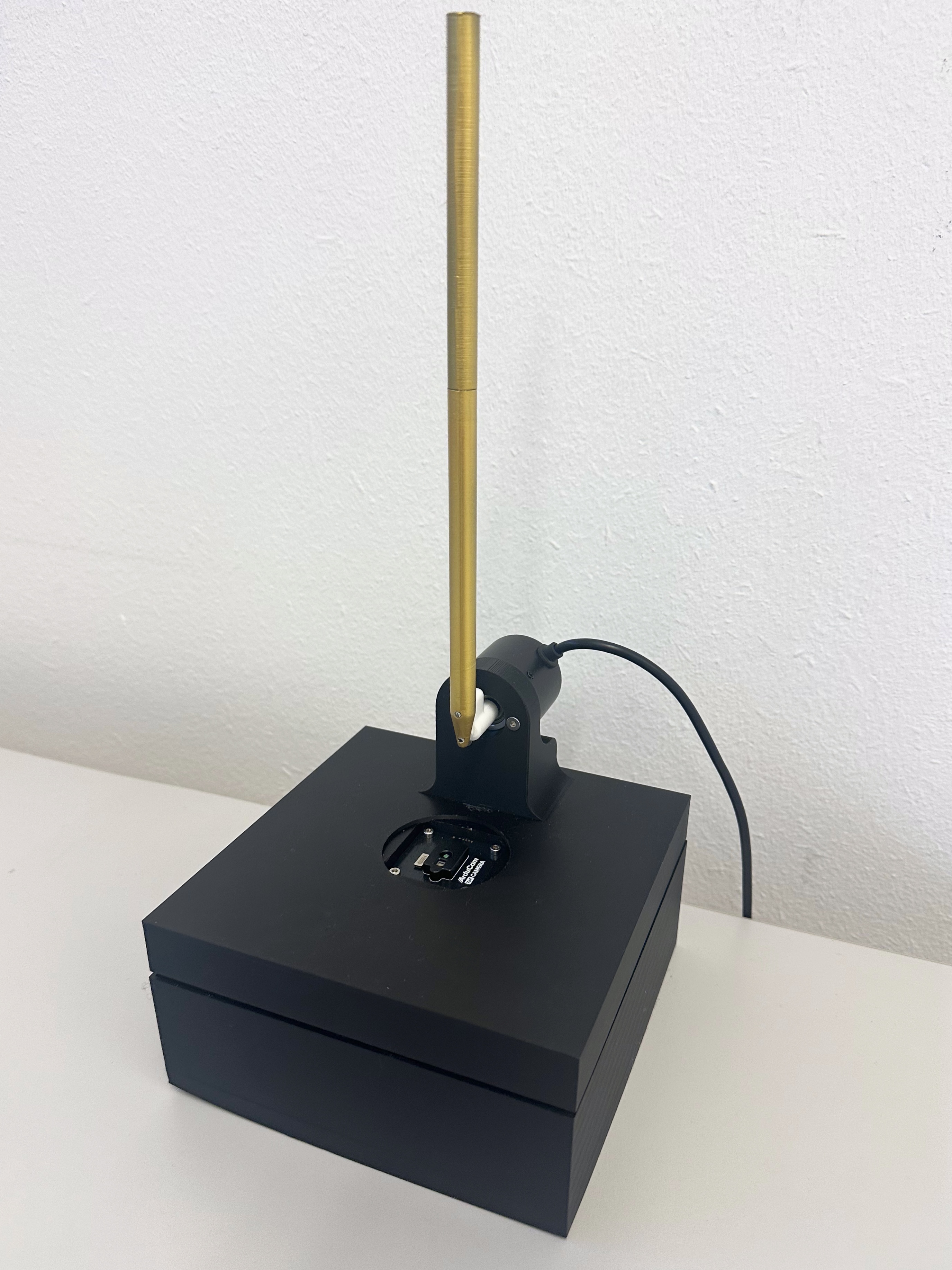}
    \caption{Our inverted pendulum (the gold pendulum standing upright) on a cart (upper part of the box) balances using angle feedback from a \gls*{tof} camera and cart measurements from the motor encoder.}
    \label{fig:futuremagicbox}
\end{figure}

\textbf{Why the cart--pole?} Prior work has explored a range of sensing and control approaches for inverted-pendulum stabilization, including low-cost vision-based methods \cite{du2020hinfopendulum,wu2022secure,wang2008stabilization,brill2016visual,fukuda2006jit,tu2011robustvisualfpga,lee2026pixels,sato2024position,hatada2022synthesis,krafes2018vision,hehn2011flying,bleher2022pixeltotorque,xu2021limits}.
Our contribution is experimental cart--pole stabilization using pendulum-angle feedback extracted from low-resolution \gls*{tof} depth images; to our knowledge, this has not been demonstrated in prior work.
Conradt et al.~\cite{conradt2009pencil} demonstrated pencil balancing using a pair of event-based dynamic vision sensors for rapid visual feedback.
Related pendulum systems have also been used to demonstrate tactile feedback for pole swing-up \cite{Bi2021ZeroShot} and magnetic actuation for spatial pendulum balancing \cite{zughaibi2025balancing}.
The upright equilibrium is open-loop unstable: without corrective cart acceleration, small perturbations grow exponentially.
\reviewchange{For the nominal acceleration-input model and parameters in \cref{sec:method,tab:params}, the unstable mode grows by a factor of $e$ in approximately $\PaperGrowthTime\,\mathrm{s}$.}
This short growth time makes timely angle feedback essential for stabilization.
Small angle errors can produce large control actions. Depth errors can alter the extracted silhouette and hence the angle feedback.
We mount the \gls*{tof} camera on the cart below the pendulum, looking upward.
This keeps the camera at a fixed position relative to the pendulum pivot.
The arrangement is analogous to eye-in-hand sensing, in which a camera moves with a robot's end-effector; \gls*{tof} cameras have been used in this configuration for target tracking \cite{reiser2007using} and robot positioning \cite{pomares2010visual}.
The task also has clear metrics, including angle error and cart oscillations.
This benchmark therefore tests whether an inexpensive, low-resolution \gls*{tof} camera can provide sufficiently accurate and timely angle feedback to stabilize an open-loop unstable equilibrium of a system with fast dynamics.

\begin{figure}[tb]
    \centering
    \begin{tikzpicture}
      \node[anchor=south west, inner sep=0] (img) at (0,0)
           {\includegraphics[width=.90\linewidth]{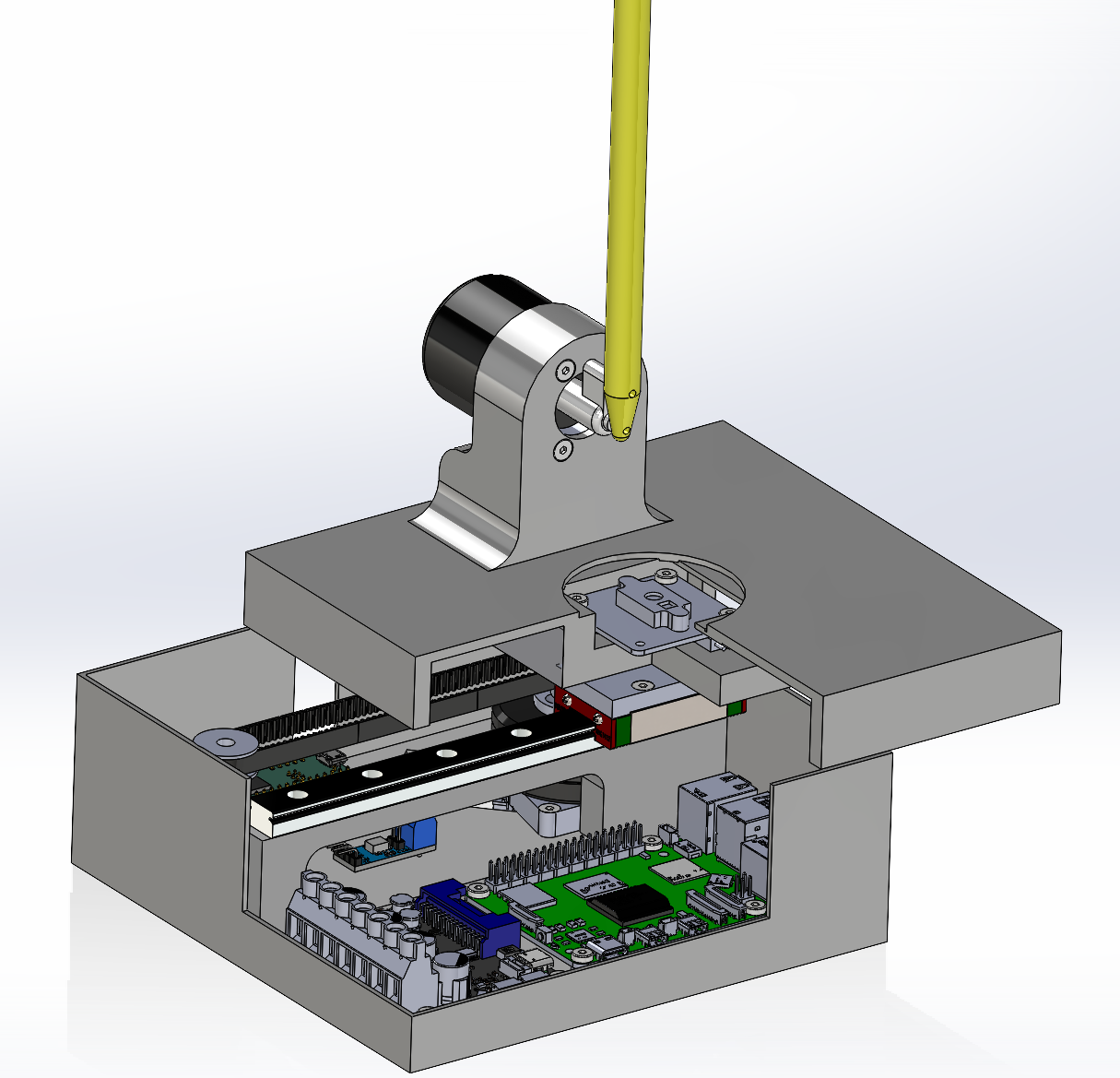}};
      \begin{scope}[reset cm, x={(img.south east)}, y={(img.north west)}]
        \tikzset{
          tag/.style={draw=red, fill=white, rounded corners=2pt,
                      minimum size=12pt, inner sep=0pt,
                      font=\bfseries\sffamily\footnotesize, align=center},
          callout/.style={-Latex, very thick, red}
        }
        \newcommand{\setupLabelTo}[5]{%
          \node[tag] (L#1) at (#2,#3) {#1};
          \draw[callout] (L#1) -- (#4,#5);
        }
        \setupLabelTo{1}{0.67}{0.9}{0.55}{0.8}
        \setupLabelTo{2}{0.857}{0.747}{0.65}{0.5}
        \setupLabelTo{3}{0.939}{0.462}{0.76}{0.25}
        \setupLabelTo{4}{0.8}{0.16}{0.56}{0.44}
        \setupLabelTo{5}{0.674}{0.086}{0.55}{0.15}
        \setupLabelTo{6}{0.495}{0.050}{0.4}{0.12}
        \setupLabelTo{7}{0.316}{0.090}{0.35}{0.22}
        \setupLabelTo{8}{0.170}{0.18}{0.457}{0.335}
        \setupLabelTo{9}{0.081}{0.28}{0.27}{0.29}
        \setupLabelTo{10}{0.063}{0.51}{0.23}{0.32}
        \setupLabelTo{11}{0.123}{0.718}{0.34}{0.29}
        \setupLabelTo{12}{0.253}{0.854}{0.52}{0.34}
        \setupLabelTo{13}{0.427}{0.924}{0.45}{0.7}
      \end{scope}
    \end{tikzpicture}
    \caption{Cutaway CAD view of the belt-driven apparatus:
    (1) pendulum; (2) enclosure cover, shown cut away; (3) enclosure base;
    (4) \gls*{tof} camera; (5) Raspberry~Pi~5; (6) ODrive~S1;
    (7) \acrfull[hyper=false]{can} transceiver;
    (8) T-Motor GB4106 \acrshort[hyper=false]{bldc} motor;
    (9) microcontroller; (10) GT2 belt and pulleys; (11) linear guideway;
    (12) carriage; and (13) pendulum holder and angle encoder.}
    \label{fig:one-dimensional}
\end{figure}

\begin{mdframed}[hidealllines=true,backgroundcolor=blue!5]
\textbf{Contributions.} We demonstrate that angle feedback from an inexpensive, low-resolution \gls*{tof} camera supports reliable and precise stabilization of an open-loop unstable equilibrium in a system with fast dynamics.
We quantify balancing accuracy across repeated trials under varying lighting conditions, characterize measurement and control timing, and compare \gls*{tof} camera-based \gls*{lqg} stabilization with encoder-feedback \gls*{lqr}.
We also present the compact platform developed for these experiments, which integrates sensing and actuation using readily available components (\cref{fig:futuremagicbox,fig:one-dimensional}).
\end{mdframed}

\section{Experimental Setup}
\label{sec:one-dimensional}

\begingroup
\widowpenalty=10000
In this section, we describe the hardware and software, formulate the system dynamics, and identify the model parameters used for estimation and control.
\par
\endgroup

\subsection{Hardware and software}
\label{sec:setup-mechanics}
To start, we briefly describe the custom and third-party components
illustrated in \cref{fig:one-dimensional}.

\begin{figure}[tb]
    \centering
    \includegraphics[width=.98\linewidth,trim={0 50bp 0 0},clip]{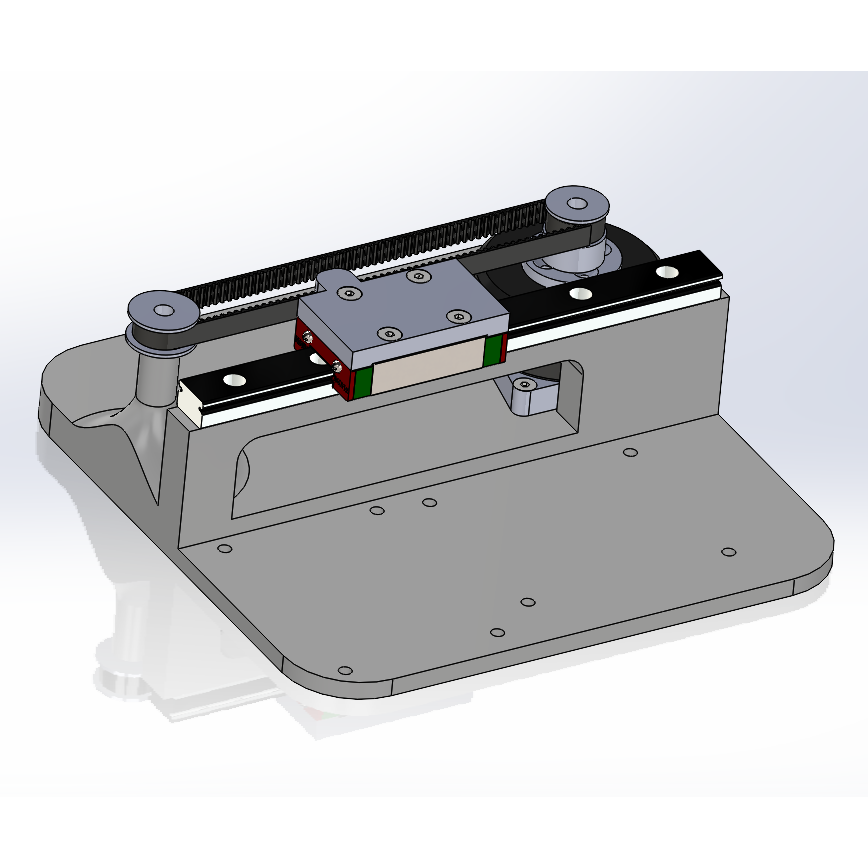}
    \par\vspace{6pt}
    \includegraphics[width=.98\linewidth,trim={0 0 0 95bp},clip]{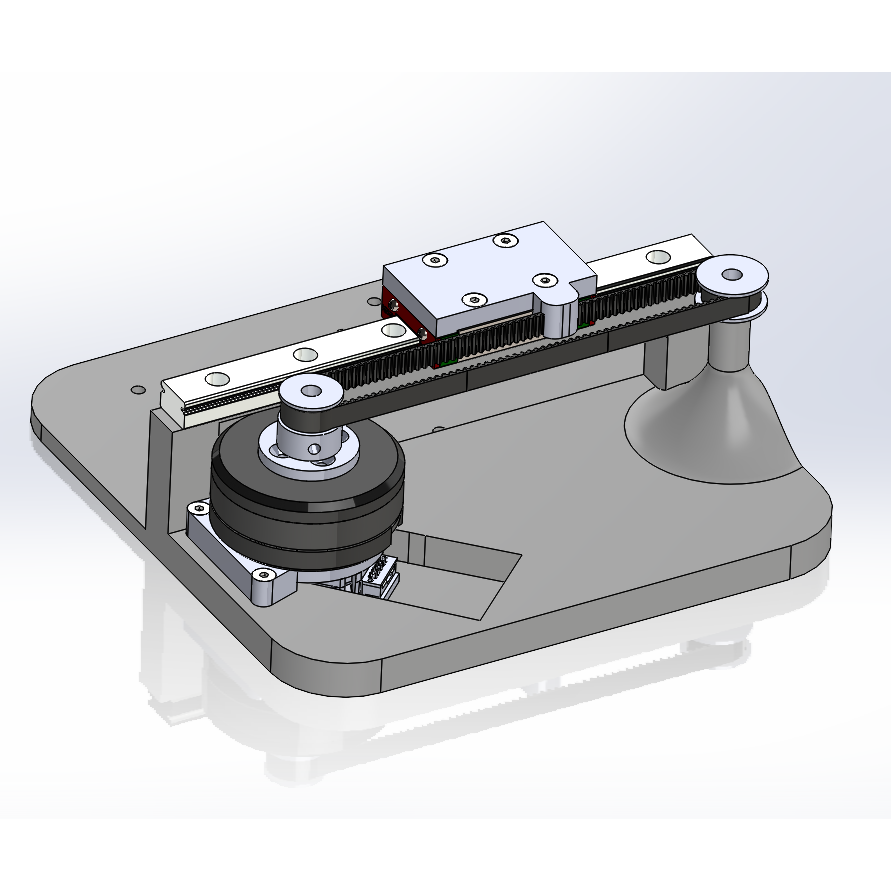}
    \caption{Belt-driven carriage on the linear guideway (top) and the
    motor and pulley arrangement (bottom), mounted on a 3D printed platform.}
    \label{fig:carriage_views}
\end{figure}

\textbf{Cart.}
The cart consists of a 3D printed platform mounted on an MGN12H carriage
and a \PaperGuidewayLengthMm{}~mm linear guideway; see \cref{fig:carriage_views}.
A T-Motor GB4106 \gls*{bldc} motor drives the cart via a \PaperBeltWidthMm{}~mm-wide
GT2 timing belt and \PaperPulleyTeeth{}-tooth pulleys.

The motor has a \PaperMotorEncoderBits{}-bit AMT212B-V-OD absolute encoder connected to an
ODrive~S1 drive over RS485.
We provide the drive with motor-velocity setpoints, which are tracked by
its velocity controller with an inner current loop.
The cart-velocity request is mapped to motor speed via
$\dot n^{\mathrm{cmd}}=v^{\mathrm{cmd}}/s_m$, where
$s_m=\PaperTransmissionScale\,\mathrm{m/rev}$ is the signed transmission scale.
The controller leaves balancing mode if $|x_c|>\PaperCartLimit\,\mathrm{m}$,
where $x_c=0$ is its initialized position.
We enforce $|v^{\mathrm{cmd}}|\leq\PaperVelocityLimit\,\mathrm{m/s}$.
These conservative cart-position and velocity-command limits
are imposed to protect the hardware.

\textbf{Pendulum.}
The pendulum is 3D printed and connected to a rotary encoder mounted
on the carriage.
As such, it is free to rotate only around a single horizontal axis;
see \cref{fig:system_1D}.
The \PaperEncoderPulsesPerRev{}-pulse-per-revolution quadrature encoder provides \PaperEncoderCounts{} counts per
revolution ($\PaperAngleBinMrad\,\mathrm{mrad}$ per count).
We denote its angle measurement by $\varphi_{\mathrm{enc}}$.
It supplies reference measurements for vision calibration and performance
evaluation, as well as initialization and safety checks.
During \gls*{tof} camera-based balancing, the pendulum-angle observations come from
the camera.

\begin{figure}[tb]
    \centering
    \includegraphics[width=.49\linewidth]{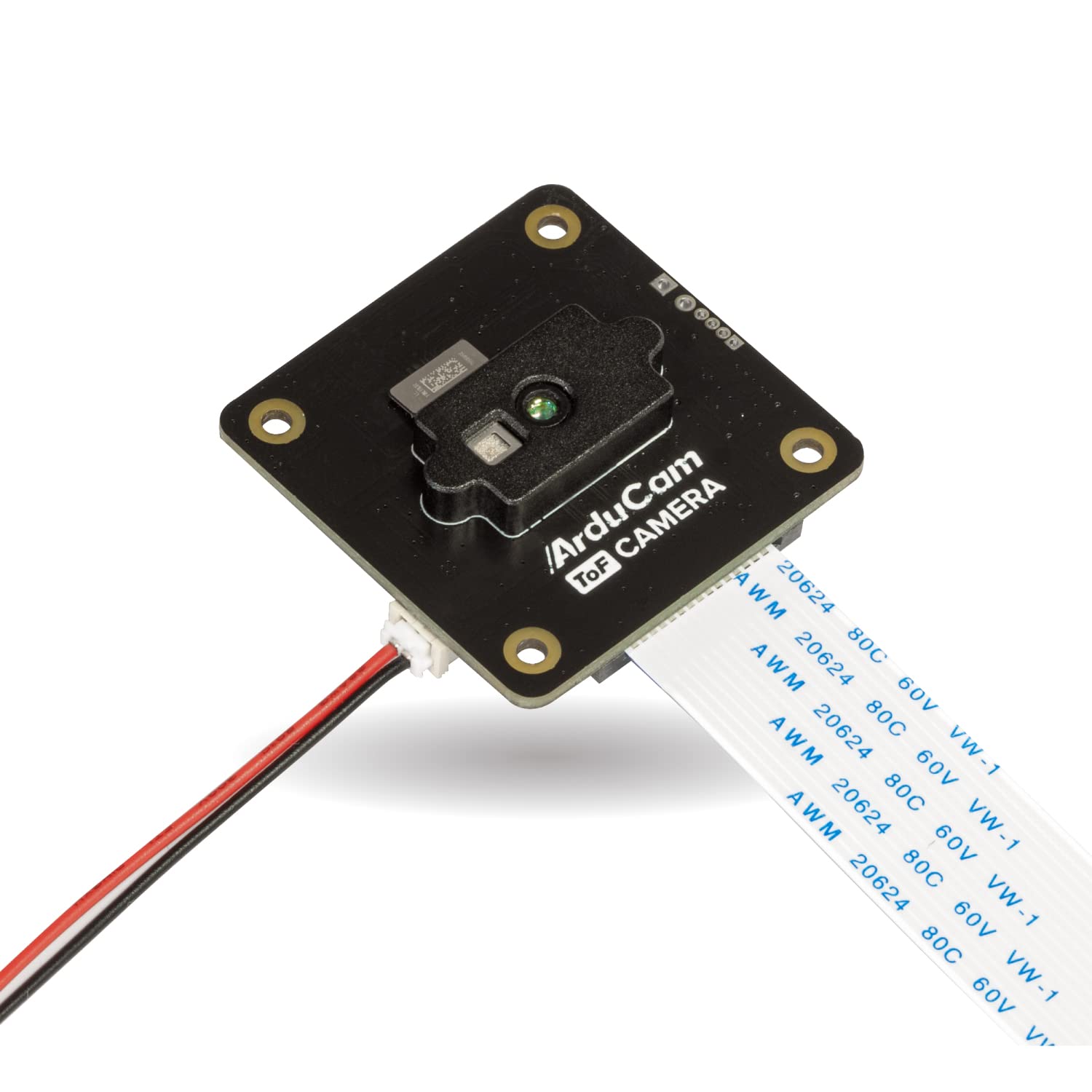}
    \hfill
    \includegraphics[width=.49\linewidth]{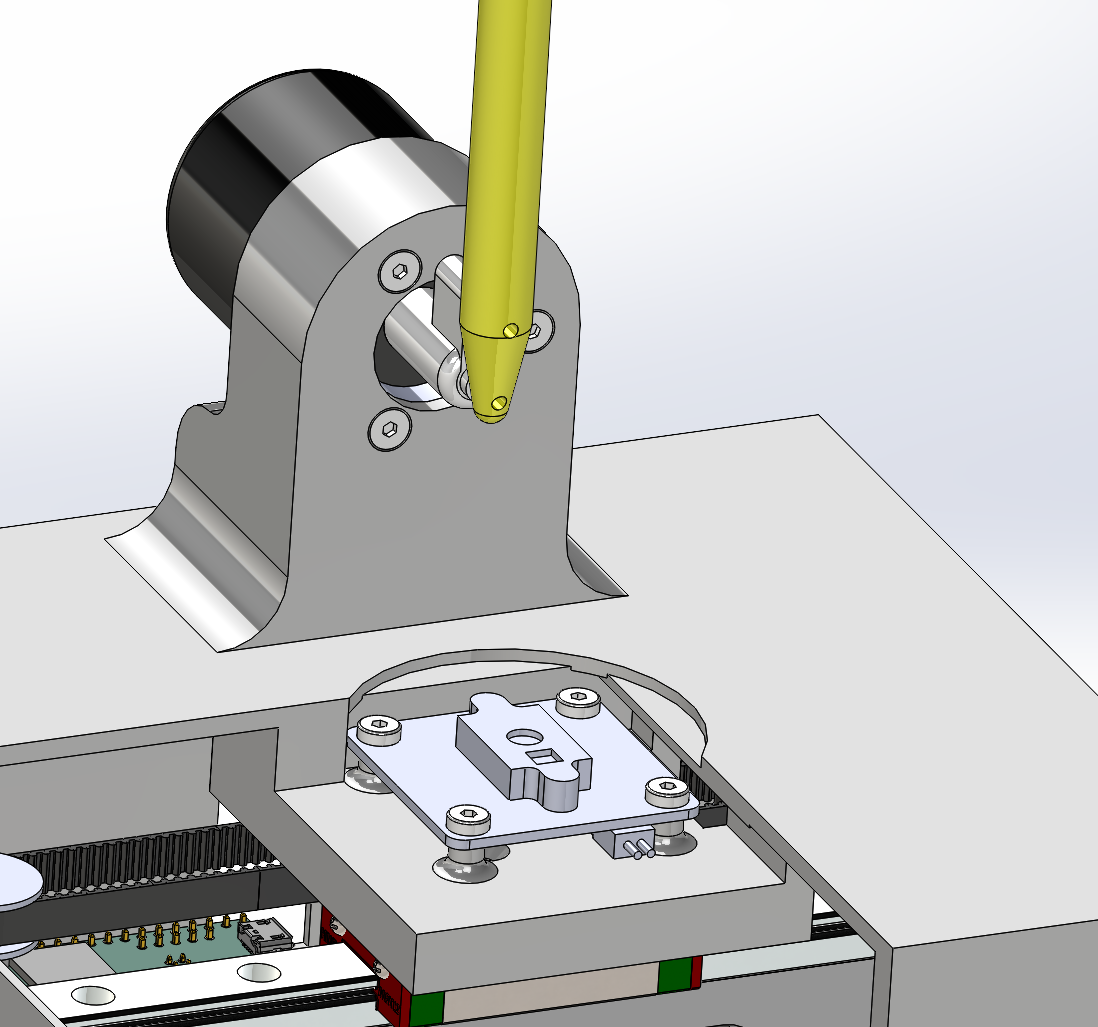}
    \caption{\gls*{tof} camera (left) and its mounting beneath the
    pendulum (right).}
    \label{fig:vision}
\end{figure}

\textbf{\gls*{tof} camera.}
We employ the Arducam \gls*{tof} camera module shown in \cref{fig:vision}.
The module operates with a \PaperCameraWavelengthNm{}~nm infrared illuminator and provides
$\PaperCameraWidth\times\PaperCameraHeight$-pixel depth frames at a nominal rate of \PaperCameraRate{}~Hz through
its CSI connection to a Raspberry~Pi~5.
Its active illumination enables depth sensing without external scene
lighting, reducing dependence on ambient visible light.
We use a custom, 3D printed holder to mount the camera directly beneath
the pendulum. \Cref{fig:alg:results} illustrates the angle extraction
described in \cref{sec:pen-measurements}.

\begin{figure}[!htb]
    \centering
    \includegraphics[width=.98\linewidth]{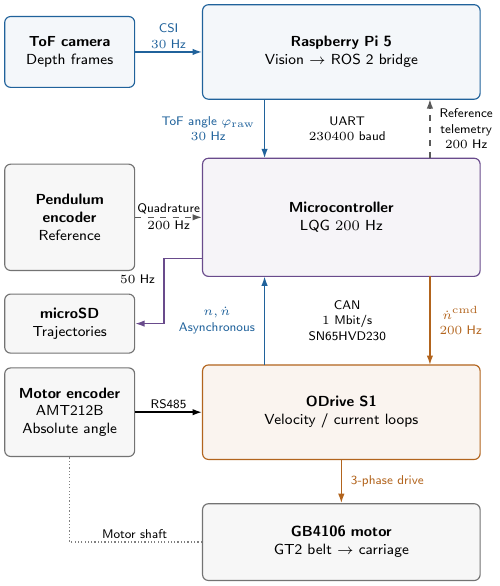}
    \caption{Data flow during \gls*{tof} camera-based balancing and trajectory logging.
    \reviewchange{Only cart position and \gls*{tof} angle supply ongoing measurement updates to the \gls*{kf};}
    dashed paths carry reference measurements.}
    \label{fig:data-flow}
\end{figure}

\textbf{Computation and communication.}
\label{sec:setup-computation}
A Teensy~4.1 microcontroller (MCU) executes the state estimation and balancing
control in C++ at \PaperControlRate{}~Hz.
The Pi runs \PaperHostNodes{} Python \gls*{ros} nodes~\cite{ros2}: one processes the
depth images, and the other forwards the estimated angle to the
microcontroller over UART at \PaperUartBaud{}~baud.
The microcontroller exchanges motor feedback and velocity commands with
the drive over CAN at \PaperCanBitrateMbps{}~Mbit/s through an SN65HVD230 transceiver.
It integrates the controller's acceleration request to obtain the velocity
setpoint and logs trajectories to microSD at \PaperLoggingRate{}~Hz.
\Cref{fig:data-flow} summarizes the signal paths.

\begin{figure}[tb]
    \centering
    \includegraphics[width=.96\linewidth]{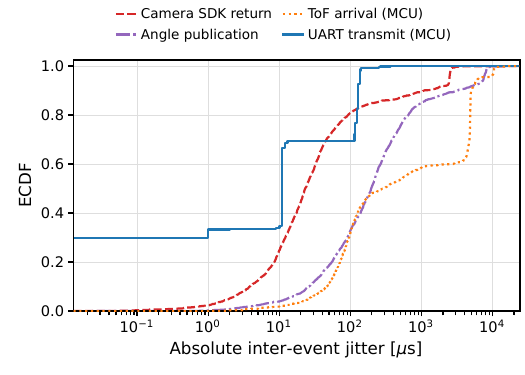}
    \caption{\Acrfullpl[hyper=false]{ecdf} of absolute inter-event jitter in $\mu$s,
    from at least \PaperTimingMinIntervals{} intervals per stream during a \PaperTimingDuration{}-second balancing run.
    Jitter is the absolute deviation of each stream's inter-event period
    from its median. The curves quantify each stream's event cadence.
    MCU \gls*{tof} arrival times are reconstructed from logged
    reception ages. MCU UART transmission carries outgoing telemetry.
    Zero-jitter samples contribute to each \gls*{ecdf} at every positive jitter threshold.}
    \label{fig:jitter}
\end{figure}

\Cref{fig:jitter} shows timing variability along the vision measurement
path and in MCU UART transmission during a \PaperTimingDuration{}-second balancing run.
For this timing capture, every MCU control tick is logged.
MCU timestamps have \PaperTimestampResolutionUs{}~$\mu$s granularity;
Pi event times are stored as integer nanoseconds from its monotonic clock.
The median control interval is \PaperTimingMedianMs{}~ms; \PaperTimingZeroPercent{}\% of the recorded control intervals
equal this median at the MCU timestamp resolution.
Camera acquisition and angle delivery occur asynchronously.

A separate \PaperTofDelayDuration{}-s capture records the encoder and raw
\gls*{tof} angles, depth frames, and Pi processing timestamps.
We divide it into \PaperTofDelayWindows{} non-overlapping
\PaperTofDelayInterval{}-s windows.
Within each window, least-squares temporal alignment of balancing data fits a constant delay
and angular offset between fresh, calibrated \gls*{tof} angles before
Kalman filtering and the linearly interpolated encoder trace, retaining
the calibrated angle scale. Camera measurements are timestamped at their
logged \gls*{kf} correction ticks. We search from
$\PaperTofDelaySearchMinMs$ to $\PaperTofDelaySearchMaxMs$~ms in
$\PaperTofDelaySearchStepMs$~ms increments, with a
$\PaperTofDelayEdgeGuard$-s exclusion at each balancing-segment boundary.
Across these windows, the effective signal delay relative to the encoder is
$\PaperTofDelayMs\pm\PaperTofDelayStdMs\,\mathrm{ms}$.
The delay and processing times are reported as mean $\pm$ sample standard
deviation.
A breakdown of the directly timed Pi processing over
\PaperCameraTimingFrames{} frames gives
$\PaperCameraSdkToWriteMeanMs\pm\PaperCameraSdkToWriteStdMs\,\mathrm{ms}$
from SDK frame return to serial-write return. This interval includes frame
copying, image publication, angle processing, and ROS dispatch.
Of this, $\PaperCameraAlgorithmMeanMs\pm\PaperCameraAlgorithmStdMs\,\mathrm{ms}$
is spent in the vision algorithm described in \cref{sec:pen-measurements},
timed from depth-frame processing through angle estimation.
Together with the balancing results in \cref{sec:experiments}, these
measurements show that the sensing pipeline provides sufficiently timely
feedback for stabilization without explicit delay compensation.

\subsection{System dynamics}
\label{sec:setup-dynamics}

\begin{figure}[tb]
    \centering
    \begin{tikzpicture}
      \node[anchor=south west, inner sep=0] (img)
        {\includegraphics[width=0.95\linewidth]{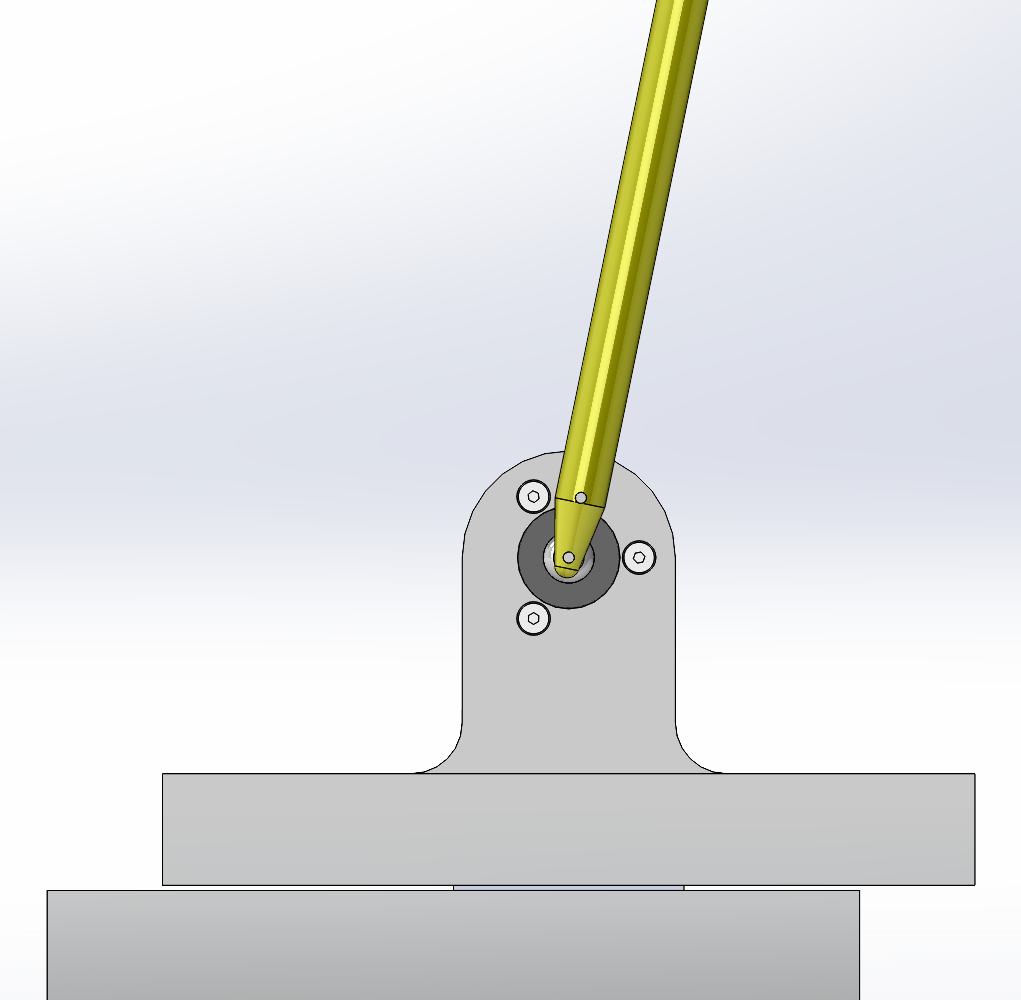}};
      \begin{scope}[x={(img.south east)}, y={(img.north west)},
                    every node/.style={font=\small}]

        \draw[-{Latex[length=3mm]}] (0.13,0.65) -- (0.13,0.42)
            node[midway,left] {$g$};

        \draw[->] (0.45,0.2) -- (0.5565,0.2) node[midway,below] {$x_c$};

        \draw[line width=1pt,-{Latex[length=3.5mm]}]
             (0.85,0.18) -- (0.999,0.18) node[midway,above] {$a$};

        \node at (0.187,0.2) {$M$};

        \node at (0.5,0.41) {$b$};

        \draw[|-|] (0.49,0.452) -- (0.583,0.95)
            node[midway,above right=1pt] {$l$};

        \node at (0.71,0.81) {$m,J$};

        \draw[->]
            (0.5565,0.35) arc[start angle=-90, end angle=75, radius=0.09];

        \node at (0.63,0.35) {$\varphi$};

        \draw[thin,black] (0.45,0.0) -- (0.45,0.23);

        \draw[thin,black] (0.5565,0.13) -- (0.5565,0.42);

      \end{scope}
    \end{tikzpicture}
    \caption{Schematic of the cart-pendulum. The angle $\varphi$ is measured
    from the downward vertical; $a$ denotes cart acceleration.}
    \label{fig:system_1D}
\end{figure}

We illustrate the schematic of our setup in \cref{fig:system_1D}.
Let $\varphi$ denote the pendulum angle relative to the downward vertical,
with upright at $\varphi=\pi$, and $x_c$ the linear cart displacement.
With cart mass $M$, pendulum mass $m$, pivot-to-center-of-mass distance
$l$, and pendulum inertia $J$ about the pivot, the classical coupled
equations of motion are
\begin{align}
    J\ddot\varphi+ml\ddot x_c\cos\varphi+mgl\sin\varphi
    &=-b\dot\varphi,
    \label{eq:phi_ddot}\\
    (M+m)\ddot x_c+ml\ddot\varphi\cos\varphi
        -ml\dot\varphi^2\sin\varphi
    &=F_{\mathrm{drv}}-f_c(\dot x_c),
    \label{eq:x_ddot}
\end{align}
\begingroup
\widowpenalty=10000
where $F_{\mathrm{drv}}$ is the horizontal drive force, $f_c$ the signed
carriage resistance, $b$ the viscous hinge damping, and $g$ gravitational
acceleration.
\par
\endgroup

\subsection{Parameter identification}
\label{sec:params}

\Cref{tab:params} lists the geometric and model parameters.
The pendulum used throughout the experiments has $l=L/2=\PaperComDistance\,\mathrm{m}$.
We use the nominal pivot inertia
$J=mL^2/12+ml^2=mL^2/3$, under the uniform-rod approximation.
We employ a batch least-squares approach to identify the hinge damping $b$.
Collected trajectories are compared to one-step predictions of the
nonlinear model using a fourth-order Runge--Kutta integrator~\cite{butcher2016numerical}.
The parameter is selected to minimize the squared prediction errors
over the collected data points using
the trust-region reflective algorithm~\cite{coleman1996interior}.

\begin{table}[tb]
    \centering
    \setlength{\tabcolsep}{4pt}
    \renewcommand{\arraystretch}{1.12}
    \begin{tabular}{lll}
    \hline
    \bfseries Parameter & \bfseries Symbol & \bfseries Value\\
    \hline
    Cart mass & $M$ & $\PaperCartMass\,\mathrm{kg}$\\
    Pendulum mass & $m$ & $\PaperPendulumMass\,\mathrm{kg}$\\
    Pendulum length & $L$ & $\PaperPendulumLength\,\mathrm{m}$\\
    COM distance & $l$ & $\PaperComDistance\,\mathrm{m}$\\
    Pivot inertia & $J$ & $\PaperPivotInertia\,\mathrm{kg\,m^2}$\\
    Hinge damping & $b$ & $\PaperHingeDamping\,\mathrm{N\,m\,s/rad}$\\
    Gravity & $g$ & $\PaperGravity\,\mathrm{m/s^2}$\\
    Signed belt scale & $s_m$ & $\PaperTransmissionScale\,\mathrm{m/rev}$\\
    \hline
    \end{tabular}
    \caption{Geometric and model parameters.}
    \label{tab:params}
\end{table}

\section{Estimation and Control}
\label{sec:method}
\reviewchange{%
In this section, we present our estimation and control pipeline, combining
a \gls*{kf} with \gls*{lqr} control~\cite{anderson2007optimal}.
We first formulate their shared model, then describe the measurements,
state estimation, and balancing controller.

To reduce reliance on explicitly modeling nonlinear carriage friction
and resistance caused by movement of the cables attached to the cart,
we use the drive's velocity-control mode.
The outer \gls*{lqr} controller produces an acceleration request $a_c$,
integrated into the velocity setpoint; the drive's $\PaperDriveControlRateKHz$~kHz internal control loop
adjusts motor current to track this setpoint.
With actual cart acceleration $a=\ddot x_c$, \eqref{eq:phi_ddot} becomes
\begin{equation}
    \ddot x_c=a,\quad
    J\ddot\varphi=-mla\cos\varphi-mgl\sin\varphi-b\dot\varphi.
    \label{eq:setup-acceleration-dynamics}
\end{equation}
The nominal model takes $a\simeq a_c$, neglecting the drive's tracking
dynamics and command saturation. Cart mass and resistance still affect
the required drive force, but do not enter the prescribed-acceleration
model; hinge damping remains.

Linearizing \eqref{eq:setup-acceleration-dynamics} around the upright
equilibrium gives $\dot z(t)=Az(t)+Bu(t)$, with cart-acceleration input
$u=a_c$ and
$z(t)=[x_c,\dot x_c,\varphi-\pi,\dot\varphi]^\top$.
Angle differences are wrapped to $[-\pi,\pi)$.
The system matrices read
\begingroup
\setlength{\arraycolsep}{4pt}
\begin{equation*}
    A=\begin{bmatrix}
        0&1&0&0\\
        0&0&0&0\\
        0&0&0&1\\
        0&0&\dfrac{mgl}{J}&-\dfrac{b}{J}
    \end{bmatrix},\qquad
    B=\begin{bmatrix}0\\1\\0\\\dfrac{ml}{J}\end{bmatrix}.
\end{equation*}
\endgroup
For controller design, we use $l=\PaperComDistance\,\mathrm{m}$ and nominal pivot inertia
$J=\PaperPivotInertia\,\mathrm{kg\,m^2}$; see \cref{tab:params}.

We discretize the continuous model with nominal sampling period $T_s=\PaperControlPeriodMs\,\mathrm{ms}$,
obtaining $A_d=e^{AT_s}$ and
$B_d=\int_0^{T_s}e^{A\tau}B\,\mathrm{d}\tau$.

Here, raw refers to measurements before processing by the \gls*{kf}; camera-angle measurements
already include image processing and calibration.
}

\subsection{Raw cart position and velocity measurements}
\label{sec:cart-measurements}
The \gls*{bldc} motor is equipped with an absolute encoder, which the drive
uses to report motor position $n$ in revolutions and velocity $\dot n$
in revolutions per second over CAN. Using the transmission scale from
\cref{sec:setup-mechanics}, the raw cart measurements are
$x_{c,\mathrm{raw}}=s_m(n-n_0)$ and
$\dot x_{c,\mathrm{raw}}=s_m\dot n$.
Here $n_0$ is the initial motor position.
\reviewchange{Cart position supplies measurement corrections to the \gls*{kf};} the reported
velocity initializes the cart-velocity estimate and command.

\subsection{Raw angle measurements}
\label{sec:pen-measurements}
\begin{figure}[tb]
    \centering
    \includegraphics[width=.49\linewidth]{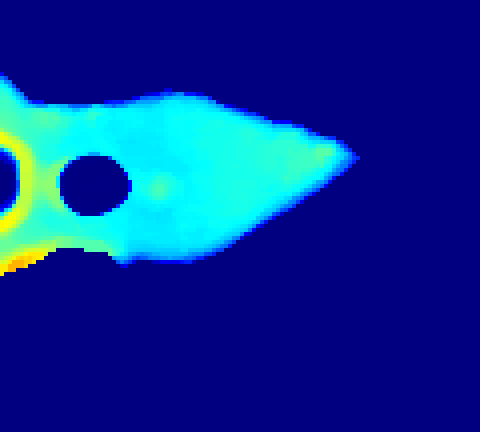}
    \hfill
    \includegraphics[width=.49\linewidth]{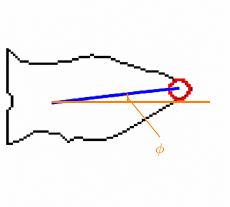}
    \caption{Offline illustration of angle extraction from a \gls*{tof} frame.
    We show the depth image after filtering (left) and the contour and
    tip circle (right). The blue segment makes angle $\phi$ with the
    orange reference and is mapped to the angle measurement $\varphi_{\mathrm{raw}}$.}
    \label{fig:alg:results}
\end{figure}

To ensure observability of the system, we complement the raw cart-position
measurement $x_{c,\mathrm{raw}}$ with angle measurements
$\varphi_{\mathrm{raw}}$. These are obtained via standard computer vision
algorithms~\cite{szeliski2022computervision} from the \gls*{tof} images,
as described in the following steps; see also \cref{fig:alg:results}.
Depth isolates the pendulum silhouette; its 2D image geometry then supplies
the angle measurement, without reconstructing a 3D pendulum axis.

\subsubsection{Image preprocessing}
We retain depths within $\PaperMinimumDepth$--$\PaperMaximumDepth\,\mathrm{m}$, map them to an image with \PaperImageBits{}-bit values,
apply a $\PaperMedianKernel\times\PaperMedianKernel$ median filter, and crop a $\PaperCropWidth\times\PaperCropHeight$-pixel region.
The largest external contour defines a filled pendulum silhouette.

\subsubsection{Locate the pendulum tip}
The pivot is fixed in image coordinates. We enlarge a padded region
around the silhouette by a factor of \PaperUpsampling{} and compute its Euclidean
distance transform. Among pixels to the right of the pivot with at least
\PaperTipClearance{} original-image pixels of boundary clearance, we select the point
farthest from the pivot. A local quadratic fit to the distance transform
refines the tip location to sub-pixel precision.

\subsubsection{Compute angle}
By connecting the fixed pivot and the center of the tip circle, we obtain
a segment with orientation $\phi$ related to the pendulum angle via
$\varphi_{\mathrm{raw}}=\alpha\phi+\beta$.
We express both angles in radians; the fitted scale $\alpha$ is dimensionless
and the fitted offset $\beta$ is in radians.
The image angle is measured from the horizontal reference, with positive
angles clockwise.
The calibration fits $\alpha$ and $\beta$ by least squares against
pendulum-encoder recordings, searching a relative time offset between
the signals and rejecting residual outliers. The pivot and tip-radius
settings remain fixed during this fit. Frames without a valid tip
produce no angle measurement.

\subsection{Sensor fusion}
\label{sec:estimate}
\reviewchange{For the \gls*{kf}, we augment the}
physical state with an additive acceleration disturbance $d_a$, modeled
as a random walk. The augmented state
$\tilde z=[z^\top,d_a]^\top$ has matrices
\begin{equation*}
    \tilde A=\begin{bmatrix}A&B\\0_{1\times4}&0\end{bmatrix},\qquad
    \tilde B=\begin{bmatrix}B\\0\end{bmatrix}.
\end{equation*}
\reviewchange{At control tick $k$, the \gls*{kf} uses the measured interval}
$T_s[k]=t[k]-t[k-1]$. The matrices $\tilde A_d[k]$ and $\tilde B_d[k]$
are obtained by the same discretization, using $\tilde A$, $\tilde B$,
and $T_s[k]$. Process noise is zero mean, with covariance $V[k]$ obtained
by integrating the continuous noise model over this interval.

\reviewchange{The known input to the \gls*{kf} is the preceding, limited velocity-command}
increment divided by the measured interval,
\begin{equation*}
    \bar u[k]=\frac{v^{\mathrm{cmd}}[k-1]-v^{\mathrm{cmd}}[k-2]}{T_s[k]}.
\end{equation*}

We assume additive, zero-mean measurement noise,
$y[k]=H\tilde z[k]+w[k]$, with diagonal covariance $W$.
The measurements are cart position and camera angle deviation from
upright: $H$ selects the first and third components of $\tilde z$.
Each fresh measurement $y_i[k]$ is processed at most once, using the
corresponding row $H_i$ and noise variance $W_{ii}$.

\reviewchange{The \gls*{kf} propagates the state estimate as}
\begin{equation*}
    \hat{\tilde z}[k]=\tilde A_d[k]\hat{\tilde z}[k-1]
                       +\tilde B_d[k]\bar u[k],
\end{equation*}
and corrects it for each available fresh measurement according to
\begin{equation*}
    \hat{\tilde z}[k]\leftarrow\hat{\tilde z}[k]
        +L_i[k]\big(y_i[k]-H_i\hat{\tilde z}[k]\big),
\end{equation*}
with the corresponding Kalman gain $L_i[k]$. The arrow denotes an in-place correction using
the pre-update estimate, one measurement at a time.
The covariance is updated in Joseph form. The reported trials use no programmed delay compensation.

At engagement, the drive and pendulum encoder initialize the four physical
states, with $\hat d_a=0$. Subsequent pendulum-angle corrections use only the camera;
the pendulum encoder remains available for safety and evaluation.

\textbf{Noise tuning.}
In a separate offline study, we used Bayesian optimization~\cite{TPE}
to search the angular process-noise densities and \gls*{tof} measurement
variance in logarithmic coordinates.
We minimized the root-mean-square prediction residual at fresh \gls*{tof} observations
before correction, using two balancing recordings for tuning and a third
for evaluation; pendulum-encoder errors served as diagnostics.
These covariances are tuning parameters and need not equal the physical
noise variances.

\subsection{Balancing control}
\label{sec:control}
\reviewchange{Using the nominal matrices $A_d$ and $B_d$ defined above, we design an
\gls*{lqr} $u[k]=-K\hat z[k]$~\cite{anderson2007optimal}
to regulate the estimated physical state to zero by minimizing the cost}
\begin{equation*}
    \sum_{k=0}^{\infty}\left(z[k]^\top Qz[k]
                    +u[k]^\top Ru[k]\right),
\end{equation*}
where $Q=\operatorname{diag}(\PaperLqrWeightPosition,\PaperLqrWeightVelocity,\PaperLqrWeightAngle,\PaperLqrWeightAngularVelocity)$ and $R=\PaperLqrInputWeight$ weight
the state deviations and acceleration input, respectively.
\reviewchange{The feedback uses the four physical components of the \gls*{kf} estimate.}
The disturbance state affects their prediction
but has no separate feedback gain.

The microcontroller integrates the acceleration request into the drive's
velocity setpoint,
\begin{equation*}
    v^{\mathrm{cmd}}[k]=\operatorname{clip}
       \left(v^{\mathrm{cmd}}[k-1]+u[k]T_s[k],-v_{\max},v_{\max}\right),
\end{equation*}
with $v_{\max}=\PaperVelocityLimit\,\mathrm{m/s}$ and the initial command set to the
measured cart velocity. No integral-error feedback is used.

\section{Experiments}
\label{sec:experiments}
In this section, we quantify pendulum-angle and cart-position errors during
repeated balancing trials with \gls*{tof} camera feedback, compare angle
regulation among indoor lighting conditions, and compare camera feedback
with encoder feedback in longer balancing runs.

\subsection{Balancing under normal room lighting}
\label{sec:experiments-balancing}
We quantify how closely the controller maintains the pendulum upright and
the cart at its reference position under normal room lighting.

\noindent\textbf{Method.}
Using the estimation and control pipeline from \cref{sec:method}, we record
\PaperBalanceTrials{} balancing trials under normal room lighting
($\approx\PaperLightingNormalIlluminance$~lux), each
$T=\PaperBalanceDuration\,\mathrm{s}$ long.
Before each trial, we manually position the pendulum near upright and
release it. Recording begins after balancing is engaged, using cart-position and camera-angle
corrections. Each trial contains \PaperBalanceSamples{} samples at \PaperLoggingRate{}~Hz.
We evaluate the pendulum-angle error $\varphi_{\mathrm{enc}}-\pi$ and cart
position $x_c$ using the pendulum and motor encoders, respectively.
All \PaperBalanceTrials{} trials are retained in the analysis.

\begin{figure}[tb]
    \centering
    \includegraphics[width=.91\linewidth]{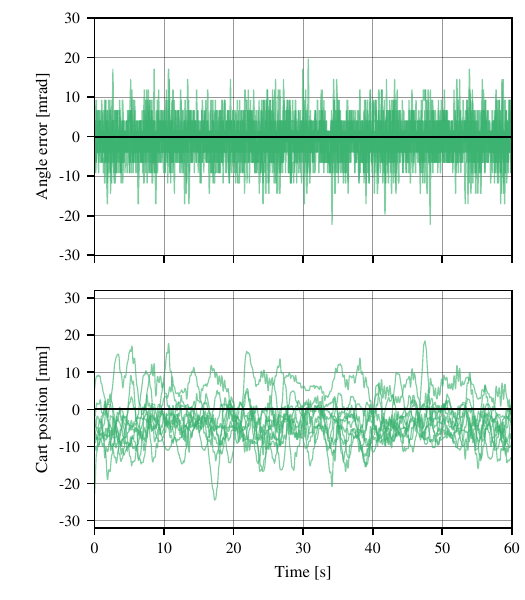}
    \caption{Pendulum-angle error $\varphi_{\mathrm{enc}}-\pi$ in mrad (top) and cart position in mm (bottom), measured by
    the pendulum and motor encoders over \PaperBalanceTrials{} successful trials
    out of \PaperBalanceTrials{} attempts, each lasting \PaperBalanceDuration{}~s, with \PaperBalanceSamples{}
    samples per trial, under normal room lighting
    ($\approx\PaperLightingNormalIlluminance$~lux). Black lines denote zero angle error and zero cart position.}
    \label{fig:control:results}
\end{figure}

\noindent\textbf{Results.}
\Cref{fig:control:results} shows the encoder-measured pendulum-angle error
and cart position across the \PaperBalanceTrials{} trials.
All \PaperBalanceTrials{} attempted trials maintained balance throughout
their \PaperBalanceDuration{}~s recordings
(\PaperBalanceTrials{}/\PaperBalanceTrials{} successful trials).
Pooled over the \PaperBalanceTrials{} trials, the angle error $\varphi_{\mathrm{enc}}-\pi$
has mean $\PaperBalanceAngleMean\,\mathrm{mrad}$ and \gls*{rmse}
$\PaperBalanceAngleRMSE\,\mathrm{mrad}$. The cart-position error relative to
$x_c=0$ has mean $\PaperBalanceCartMeanMm\,\mathrm{mm}$ and \gls*{rmse} $\PaperBalanceCartRMSEmm\,\mathrm{mm}$.
All retained samples report balancing mode. The pooled root-mean-square
discrepancies from the simultaneous encoder readings
are $\PaperBalanceRawDisagreement\,\mathrm{mrad}$ for the held raw camera angle and
$\PaperBalanceFilteredDisagreement\,\mathrm{mrad}$ for the filtered angle. These comparisons
include sampling, latency, and encoder quantization effects.

\subsection{Balancing under different lighting conditions}
\label{sec:experiments-lighting}
We examine whether balancing performance shows the expected insensitivity to changes in visible indoor lighting.

\noindent\textbf{Method.}
We collect \PaperLightingTotalTrials{} trials under \PaperLightingGroups{} indoor lighting
conditions: normal room lighting ($\approx\PaperLightingNormalIlluminance$~lux),
strong lamp illumination ($\approx\PaperLightingLampIlluminance$~lux),
and a dark room ($\approx\PaperLightingDarkIlluminance$~lux),
with \PaperLightingTrials{} consecutive trials per condition.
The normal-lighting trials are those reported in \cref{sec:experiments-balancing};
we use the same procedure for the other conditions.
For strong lamp illumination, we use a Rollei Lumis Key-Light panel
positioned approximately \PaperLightingLampDistanceCm{}~cm from and parallel
to the cart.
For each condition, we compute the mean trial-level pendulum-angle
\gls*{rmse} and a \PaperConfidencePercent{}\% Student-$t$ \gls*{ci} across trials.
We also compare normalized histograms of the pooled signed errors using Jensen--Shannon
divergence, $\frac12 D_{\mathrm{KL}}(p\|m)+\frac12 D_{\mathrm{KL}}(q\|m)$,
where $D_{\mathrm{KL}}$ is Kullback--Leibler divergence, $m=(p+q)/2$,
and logarithms have base $\PaperJsdLogBase$.
This divergence ranges from zero for identical histograms to one bit for
disjoint supports. We retain all offsets and samples, using common bins of
width $\PaperJsdAngleBin\,\mathrm{mrad}$ for angle and
$\PaperJsdCartBin\,\mathrm{mm}$ for position, with an edge offset of
$\PaperJsdEdgeOffset$ bin widths from zero.

\noindent\textbf{Results.}
\Cref{tab:lighting} summarizes the results for each condition.
Mean trial-level angle \gls*{rmse} ranges from $\PaperLightingMinRMSE$ to $\PaperLightingMaxRMSE\,\mathrm{mrad}$,
supporting stable regulation under the \PaperLightingGroups{} tested conditions.
Pairwise divergences between lighting conditions range from
$\PaperLightingAngleJsdMin$ to $\PaperLightingAngleJsdMax$ bits for angle and
$\PaperLightingCartJsdMin$ to $\PaperLightingCartJsdMax$ bits for position.
The small angle-error divergences and similar \gls*{rmse} values confirm
the expected low sensitivity to the tested visible-light conditions, consistent
with the camera's active \PaperCameraWavelengthNm{}~nm illumination.

\begin{table}[tb]
    \centering
    \caption{Pendulum-angle regulation under \PaperLightingGroups{} indoor lighting
    conditions. Means and \PaperConfidencePercent{}\% Student-$t$ \glspl*{ci} are computed across
    per-trial \gls*{rmse} values.}
    \label{tab:lighting}
    \small
    \setlength{\tabcolsep}{3pt}
    \begin{tabular}{lcrr}
        \hline
        Condition & Trials & Mean \gls*{rmse} & \PaperConfidencePercent{}\% \gls*{ci} \\
                  &        & [mrad] & [mrad] \\
        \hline
        \PaperLightingTableRows
        \hline
    \end{tabular}
\end{table}

\subsection{Comparison with encoder feedback}
\label{sec:experiments-encoder-comparison}
We compare the distributions of pendulum-angle and cart-position errors
during sustained balancing with camera feedback and encoder feedback.

\noindent\textbf{Method.}
We compare \gls*{tof} camera-based \gls*{lqg} with a conventional encoder-feedback \gls*{lqr}
using the same gains and \PaperControlRate{}~Hz control loop. In the baseline, the \gls*{lqr}
uses drive-reported cart position and velocity, pendulum encoder angle,
and angular velocity computed by finite differences of that angle at each
control tick. We bypass the \gls*{kf} because the encoders provide accurate
reference measurements at the control-loop rate with negligible acquisition
latency. This gives the encoder-derived state full weight, analogous to a
Kalman update with direct state measurements and vanishing measurement-noise
covariance.
For each configuration, we record one
\PaperComparisonDuration{}~s trajectory at \PaperLoggingRate{}~Hz after
balancing is established. Both runs use normal room lighting
($\approx\PaperLightingNormalIlluminance$~lux).
We evaluate both runs using the pendulum encoder
angle and motor-encoder cart position.
We compute Jensen--Shannon divergence with the same histogram estimator as
in \cref{sec:experiments-lighting}. To assess binning sensitivity, we vary
width multipliers over $\{\PaperJsdWidthMultipliers\}$ and edge offsets over
$\{\PaperJsdEdgeOffsets\}$ bin widths.

\begin{figure}[tb]
    \centering
    \includegraphics[width=.91\linewidth]{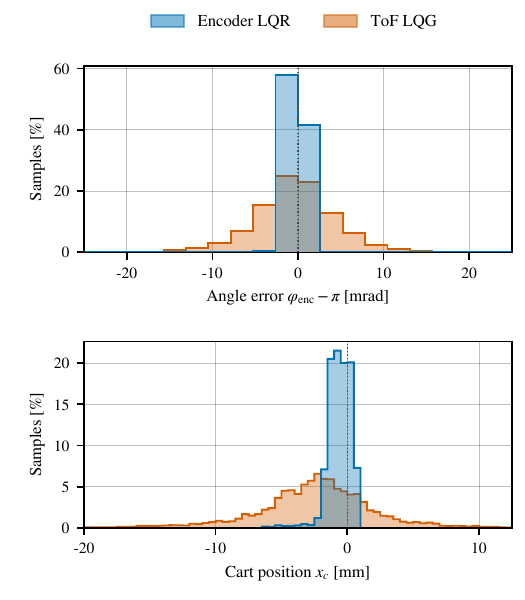}
    \caption{Empirical distributions of pendulum-angle error in mrad (top)
    and cart-position error in mm (bottom) during sustained balancing,
    measured by the pendulum and motor encoders, respectively.
    \Gls*{tof} \gls*{lqg} and encoder \gls*{lqr} each contribute \PaperComparisonSamples{} samples from one
    \PaperComparisonDuration{}~s trial. Shaded histograms use common bins of width
    $\PaperAngleBinMrad\,\mathrm{mrad}$ (one encoder count) and
    $\PaperCartBinMm\,\mathrm{mm}$, respectively. Recorded offsets are retained.
    Axes are zoomed; percentages use all samples.}
    \label{fig:control:feedback-comparison}
\end{figure}

\noindent\textbf{Results.}
\Cref{fig:control:feedback-comparison} compares the empirical distributions
of the encoder-measured angle and cart-position errors in both runs.
The encoder-feedback baseline gives angle and cart \glspl*{rmse} of
$\PaperComparisonEncoderAngleRMSE\,\mathrm{mrad}$ and $\PaperComparisonEncoderCartRMSE\,\mathrm{mm}$, respectively,
compared with $\PaperComparisonTofAngleRMSE\,\mathrm{mrad}$ and $\PaperComparisonTofCartRMSE\,\mathrm{mm}$
for \gls*{tof} camera-based \gls*{lqg}. Mean angle and cart errors are
$\PaperComparisonEncoderAngleMean\,\mathrm{mrad}$ and $\PaperComparisonEncoderCartMean\,\mathrm{mm}$ for
encoder \gls*{lqr}, and $\PaperComparisonTofAngleMean\,\mathrm{mrad}$ and
$\PaperComparisonTofCartMean\,\mathrm{mm}$ for \gls*{tof} \gls*{lqg}.
Residual angle-calibration error may contribute to these offsets,
since the position and angle feedback terms can balance at a nonzero cart
position in the absence of integral-error feedback.
Future work will augment the \gls*{kf} with a varying
angle-bias state to estimate and compensate for residual calibration error
online.
The angle and position divergences are $\PaperComparisonAngleJsd$ and
$\PaperComparisonCartJsd$ bits, respectively. Varying bin width and alignment
gives ranges of $\PaperComparisonAngleJsdSensitivityMin$--$\PaperComparisonAngleJsdSensitivityMax$
and $\PaperComparisonCartJsdSensitivityMin$--$\PaperComparisonCartJsdSensitivityMax$ bits;
the angle comparison is particularly sensitive to histogram resolution.
The lower \glspl*{rmse} and narrower error distributions indicate tighter
angle and cart-position regulation with encoder feedback.
Both configurations maintain balance throughout their
\PaperComparisonDuration{}~s recordings, demonstrating that \gls*{tof} camera
feedback also supports sustained upright regulation.

\section{Conclusions}
\label{sec:conclusion}
In this paper, we revisit a canonical control benchmark with a practical question in mind: can an inexpensive, low-resolution \gls*{tof} camera deliver pendulum-angle measurements suitable for precise feedback control? We provide a \emph{proof of existence} that it can. Beyond this empirical demonstration, we present a compact, fully enclosed platform that integrates sensing and actuation using readily available components. Importantly, the reported performance should be viewed as a \emph{lower bound} on what is achievable with \gls*{tof}-based feedback: future work can explore alternative \gls*{tof} sensors, placements, and multi-view configurations, and pair them with more advanced control design---such as model-predictive control \cite{richalet1978mphc,rawlings2026mpc} or hard-constrained neural network controllers \cite{grontas2026pinetoptimizinghardconstrainedneural} to explicitly enforce constraints and target metrics such as reduced settling time, or robust \cite{zhou1996robust} and distributionally robust controllers \cite{lanzetti2026optimalitylinearpoliciesdistributionally} to address uncertainty in plant dynamics and noise distributions. We hope this combination of evidence and tooling helps shift attention from ``which premium sensor?'' to ``how far can thoughtful sensing and system design take us?''

\bibliographystyle{IEEEtran}
\bibliography{main}

\end{document}